\documentclass[journal]{IEEEtran}
\IEEEoverridecommandlockouts

\usepackage{cite}
\usepackage{amsmath,amssymb,amsfonts}
\usepackage{graphicx}
\usepackage{textcomp}
\usepackage{xcolor}
\usepackage{tabularx}
\usepackage{diagbox}
\usepackage{siunitx}
\usepackage{adjustbox}
\usepackage{siunitx}
\usepackage{multirow}
\usepackage{booktabs}
\usepackage{array}
\usepackage{multirow}
\usepackage{setspace} 
\usepackage{arydshln}
\usepackage{algorithmicx}
\usepackage{algcompatible}
\usepackage{algorithm}
\usepackage{xcolor}
\usepackage{bm}
\usepackage{graphicx}
\usepackage{subcaption}
\usepackage{booktabs}
\usepackage{multirow} 
\usepackage{threeparttable}  
\usepackage{booktabs}
\usepackage{siunitx}
\usepackage{pifont}
\newcommand{\cmark}{\ding{51}}
\newcommand{\xmark}{\ding{55}}
\usepackage{caption}
\usepackage{booktabs}
\usepackage{multirow}
\usepackage{array}

\usepackage{algpseudocode}

\begin{document}
\title{Multi-Joint Physics-Informed Deep Learning Framework for Time-Efficient Inverse Dynamics
\\
\author{Shuhao Ma, 
        Zeyi Huang,
        Yu Cao, ~\IEEEmembership{Member, IEEE},
        Wesley Doorsamy,~\IEEEmembership{Senior Member, IEEE},
        Chaoyang Shi,~\IEEEmembership{Senior Member, IEEE},
        Jun Li,
        Zhi-Qiang Zhang,~\IEEEmembership{Senior Member, IEEE}}

\thanks{This work was supported in part by the China Scholarship Council (CSC) under Grant 202208320117. (Corresponding author: Zhi-Qiang Zhang)}
\thanks{Shuhao Ma, Yu Cao, Wesley Doorsamy, and Zhi-Qiang Zhang are with the School of Electronic and Electrical Engineering, University of Leeds, Leeds LS2 9JT, U.K. (e-mail: elsma@leeds.ac.uk; y.cao1@leeds.ac.uk; W.Doorsamy@leeds.ac.uk; z.zhang3@leeds.ac.uk).}
\thanks{Chaoyang Shi and Zeyi Huang are with the Key Laboratory of Mechanism Theory and Equipment Design of the Ministry of Education, School of Mechanical Engineering, Tianjin University, Tianjin 300072, China. (e-mail: chaoyang.shi@tju.edu.cn; zeyi.huang@tju.edu.cn).}
\thanks{Jun Li is with College of Intelligent Systems Science and Engineering, Hubei Minzu University, 39 Xueyuan Road, Enshi, 445000, Hubei, China. (e-mail:1995007@hbmzu.edu.cn)}
}
\maketitle

\begin{abstract}
Time-efficient estimation of muscle activations and forces across multi-joint systems is critical for clinical assessment and assistive device control.
However, conventional approaches are computationally expensive and lack a high-quality labeled dataset for multi-joint applications.
To address these challenges, we propose a physics-informed deep learning framework that estimates muscle activations and forces directly from kinematics. The framework employs a novel Multi-Joint Cross-Attention (MJCA) module with Bidirectional Gated Recurrent Unit (BiGRU) layers to capture inter-joint coordination, enabling each joint to adaptively integrate motion information from others. By embedding multi-joint dynamics, inter-joint coupling, and external force interactions into the loss function, our Physics-Informed MJCA-BiGRU (PI-MJCA-BiGRU) delivers physiologically consistent predictions without labeled data while enabling time-efficient inference.
Experimental validation on two datasets demonstrates that PI-MJCA-BiGRU achieves performance comparable to conventional supervised methods without requiring ground-truth labels, while the MJCA module significantly enhances inter-joint coordination modeling compared to other baseline architectures.
\end{abstract}


\IEEEpeerreviewmaketitle

\section{Introduction}\label{intro}
Multi-joint coordination in human movement represents one of the most complex motor control tasks in biomechanics, requiring precise temporal and spatial synchronization of multiple muscle groups across interconnected joints and degrees of freedom (DOFs)\cite{Dubois-JoNB-2023, Liang-TNSRE-2023}. Accurate estimation of muscle activation patterns and force distributions across the multi-joint, multi-degree-of-freedom (DOF) systems is essential for clinical movement assessment \cite{Wang-TNSRE-2025}, assistive device control \cite{Zhang-TASE-2025, cao-Tmech-2025, Cao-TRO-2025}, injury prevention \cite{Collings-MSSE-2023}, and neuromotor disorder diagnosis \cite{Makaram-TIM-2021}. 
However, direct measurement of these muscle forces and activations presents significant technical challenges, particularly for deep-seated muscles spanning multiple joints that are inaccessible to surface electromyography (sEMG) \cite{TangTNSRE2021, HuTNSRE2022, RobinTNSRE2023}.

To address these measurement limitations, traditional musculoskeletal (MSK) modeling approaches estimate muscle states from observable joint kinematic data. 
Dynamic optimization (DO) ensures temporal consistency by integrating muscle dynamics across the entire movement, but requires hours to days of computation \cite{Anderson-JoBE-2001}.
To overcome this computational burden, static optimization (SO) simplifies the problem by decoupling temporal dependencies and solving each time frame independently \cite{Ao-JoNB-2024, Li-TNSRE-2022}.
At each time step, SO estimates muscle activations by enforcing joint torque equilibrium while minimizing a physiological cost function, such as squared activations or metabolic energy \cite{zajac-CRiBE-1989}.
Despite sacrificing temporal consistency, this decoupled approach reduces computational complexity by orders of magnitude, enabling its widespread adoption in multi-joint applications.
For instance, 
Anderson et al. applied SO to estimate activations for 54 lower limb muscles during walking \cite{Anderson-JoBE-2001}. 
Hamner et al. utilized SO to analyze muscle contributions across hip, knee, and ankle joints during accelerated running \cite{HAMNER-JoB-2010}. 
However, SO becomes increasingly computationally expensive in multi-joint systems, where muscle redundancy further increases the problem complexity \cite{VALERO-JoB-2015, Sobinov-PLOs-2020, Cohn-FiRS-2023}.
Consequently, SO requires seconds to minutes per movement sequence—far exceeding the 75 ms threshold for real-time biofeedback \cite{Kannape-JoN-2013}, rendering it unsuitable for real-time applications \cite{TRINLER-JoB-2019}.

In response to the computational burden of these traditional MSK inverse dynamics optimization approaches, data-driven methods, particularly deep learning, have emerged as promising alternatives for muscle states estimation \cite{Nasr-FiCN-2021, Schmidt-BEO-2023}. These models bypass iterative optimization by directly learning the nonlinear mapping from joint kinematics to muscle states, offering near-instantaneous inference suitable for real-time applications \cite{Rane-AoBE-2019}. Recurrent architectures are among the most commonly used approaches: for example, Schmidt et al. predicted upper-limb activations with an LSTM network \cite{Schmidt-BEO-2023}, and Khant et al. estimated lower-limb activations during gait using recurrent models \cite{khant-MDPIS-2023}. 
More recently, attention-based architectures such as transformers have demonstrated superior ability to capture long-range dependencies and coordination patterns in movement-related tasks \cite{wei-jbhi-2025, lin-TNSRE-2023}, suggesting strong potential for modeling inter-joint coupling in multi-joint musculoskeletal systems.
Despite these advances, purely data-driven approaches face two critical limitations. They require large, high-quality labeled datasets that are costly and difficult to obtain at scale \cite{Horst-SD-2021, Falisse-TBME-2025}, and as unconstrained “black boxes,” they lack biomechanical guidance and may yield physiologically implausible predictions in multi-joint contexts \cite{Karniadakis-NRP-2021}. 

The limitations of both traditional MSK modeling and purely data-driven methods have motivated the development of hybrid approaches.
By embedding biomechanical principles into neural network frameworks, these models aim to regularize predictions, penalize non-physiological outputs, and enhance interpretability \cite{Zhang-TNSRE-2023, Zhang-JBHI-2022, Han-TIM-2025}. 
Despite this progress, the inverse dynamics tasks of estimating muscle activations and forces from kinematics remains underexplored, as biomechanical constraints alone cannot resolve muscle redundancy without additional performance criteria embedded in training. Multi-joint systems further complicate this through inter-joint coupling, bi-articular contributions, and exponentially growing solution spaces.
Our previous work initiated a step in this direction \cite{Ma-TNSRE-2025}, yet it could not address the challenge of multi-joint coordination where muscle activations and forces depend on the coupled dynamics of the entire kinematic chain.
Therefore, this paper introduces a physics-informed Multi-Joint Cross-Attention with Bidirectional Gated Recurrent Unit (PI-MJCA-BiGRU), to enable accurate estimation of muscle activations and forces in complex multi-joint systems.
The main contributions of this work include:
1) \textbf{Physics-informed multi-joint dynamics integration.} 
We extend the physics-informed deep learning framework to multi-joint systems by integrating complex multi-joint MSK dynamics and inter-joint coupling dynamics (including external forces) into the loss function, enabling label-free training while ensuring biomechanical consistency.
2) \textbf{Joint cross-attention architecture for coordination modeling.}
We develop MJCA-BiGRU, a novel architecture specialized in learning inter-joint coupling representations from multi-joint kinematic data.
The core of MJCA-BiGRU is a joint cross-attention mechanism that enables each joint to selectively attend to the motion states of other joints, effectively capturing complex coordination patterns.
Experiments demonstrate that the proposed framework, trained solely with physics-informed loss and no ground-truth labels, achieves representations and performance comparable to supervised learning, with ablation studies confirming MJCA's superior feature extraction over baseline variants.

The remainder of this paper is organized as follows: The configuration of our proposed method and the design of loss functions are introduced in Section \ref{method}. 
Dataset and experimental preparation are then described in Section \ref{Dataset}. 
Section \ref{results} reports the experimental results of two datasets, 
and discussions about the limitations and future research avenues are presented in Section \ref{discussions}. Section \ref{conclusion} concludes this article.

\section{Methods}\label{method}
\subsection{Main Framework}
\begin{figure*}
  \centering
  \includegraphics[scale = 0.65]{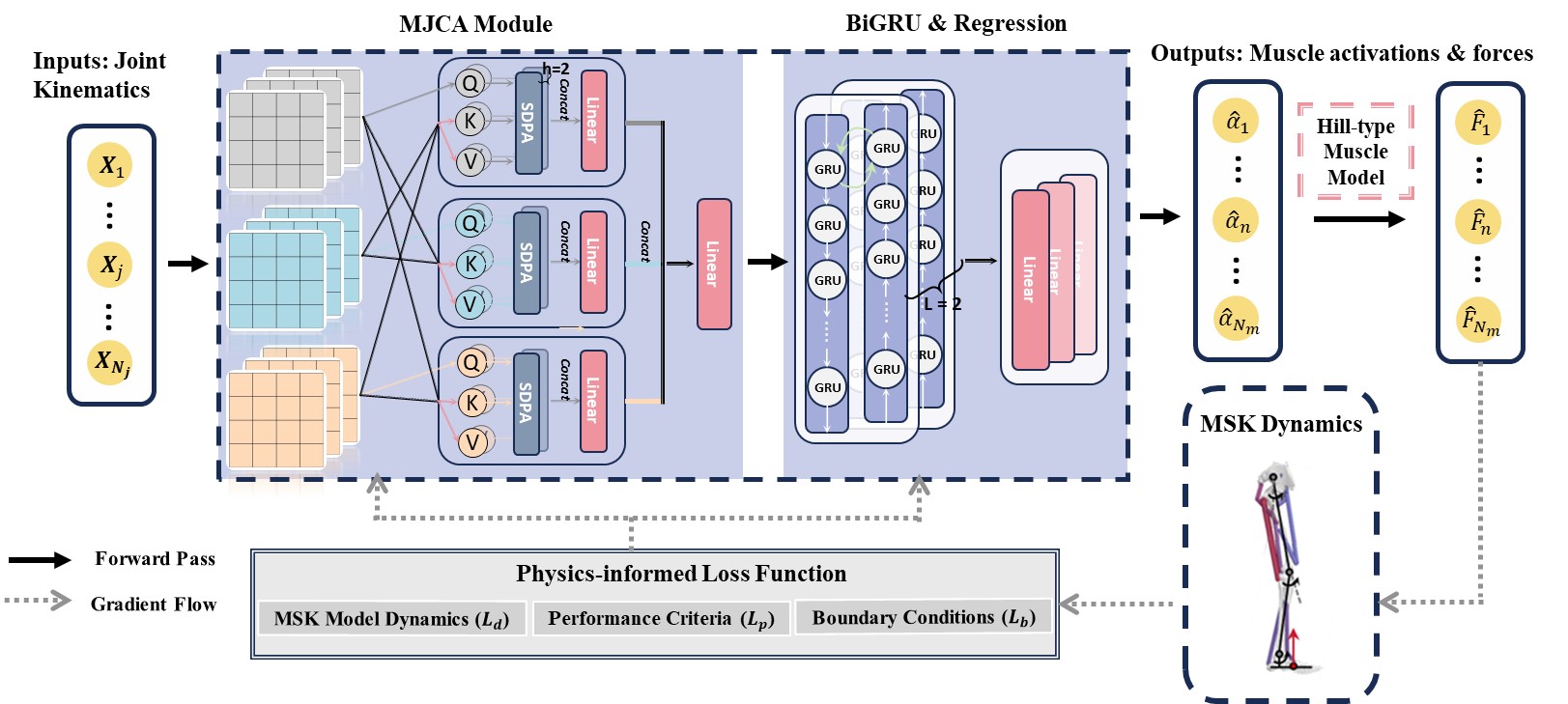} 
  \caption{Overview of the proposed PI-MJCA-BiGRU framework.}
  \label{mainframework}
\end{figure*}
Fig.~\ref{mainframework} illustrates the architecture of the PI-MJCA-BiGRU framework, which directly estimates muscle activations and forces from multi-joint kinematic data. Specifically, the MJCA-BiGRU network takes as inputs the joint kinematics $\mathbf{X} = (\mathbf{X}_1, \ldots, \mathbf{X}_j,\ldots, \mathbf{X}_{N_j})$ for $N_j$ joints, where each $X_j$ contains the $j$-th joint's angles $q_j$, angular velocities $\dot{q}_j$, and angular accelerations $\ddot{q}_j$. The input kinematic data are first decomposed into joint-specific representations, passed through an MJCA module to capture inter-joint dependencies, integrated across joints, modeled temporally by a BiGRU, and finally mapped to muscle activations through a fully connected network (FCN).
The network outputs muscle activations $\hat{\mathbf{a}} = (\hat{a}_1, \ldots, \hat{a}_{N_m})$ for $N_m$ muscles, which are then fed into an embedded hill-type muscle model to compute the corresponding forces $\hat{\mathbf{F}} = (\hat{F}_1, \ldots, \hat{F}_{N_m})$.
The framework employs a physics-informed loss function comprising three components: MSK model dynamics ($\mathcal{L}_d$), performance criteria ($\mathcal{L}_p$), and boundary conditions ($\mathcal{L}_b$).

\subsection{Design of MJCA-BiGRU}
\subsubsection{Joint-Specific Feature Extraction}
For each joint $j$, the kinematic sequence consists of angle, velocity, and acceleration:
\begin{equation}
\mathbf{X}_j = [q_j, \dot{q}_j, \ddot{q}_j] \in \mathbb{R}^{B \times S \times 3}, \quad j \in \{1,...,N_j\},
\end{equation}
where $B$ is the batch size and $S$ is the sequence length. Each sequence is projected into a hidden dimension $D_l$ through a linear layer with GELU activation:
\begin{equation}
\mathbf{H}_j = \text{GELU}(\text{Linear}(\mathbf{X}_j)) \in \mathbb{R}^{B \times S \times D_l}.
\end{equation}

\subsubsection{MJCA}
The MJCA module enables each joint to attend to all other joints. For joint $j$, the cross-joint feature matrix is constructed by concatenating other joints’ features along the feature dimension:
\begin{equation}
\mathbf{C}_j = \text{Concat}_{k \neq j}(\mathbf{H}_k) \in \mathbb{R}^{B \times S \times (N_j-1)D_l}.
\end{equation}
From $\mathbf{H}_j$ and $\mathbf{C}_j$, queries, keys, and values are obtained via learned linear projections:
\begin{align}
\mathbf{Q}_j &= \mathbf{H}_j\mathbf{W}^Q_j, \\
\mathbf{K}_j &= \mathbf{C}_j\mathbf{W}^K_j, \\
\mathbf{V}_j &= \mathbf{C}_j\mathbf{W}^V_j,
\end{align}
where $\mathbf{W}^Q_j, \mathbf{W}^K_j, \mathbf{W}^V_j \in \mathbb{R}^{D_l \times D_l}$.
These are split into $N_h$ heads with dimension $d_k = D_l/N_h$. For each head $h$, attention is computed as
\begin{equation}
\mathbf{A}^{(h)}_j = \text{softmax}\!\left(\frac{\mathbf{Q}^{(h)}_j(\mathbf{K}^{(h)}_j)^T}{\sqrt{d_k}}\right)\mathbf{V}^{(h)}_j.
\end{equation}
Outputs from all heads are concatenated, projected back to $\mathbb{R}^{B \times S \times D_l}$, and combined with the original representation by a residual connection:
\begin{equation}
\mathbf{H}'_j = \mathbf{A}_j + \mathbf{H}_j.
\end{equation}
Applied in parallel to all joints, this yields updated features $\{\mathbf{H}'_j\}$ that encode both joint-specific and inter-joint information.

\subsubsection{Joint Feature Integration}
The enhanced joint features are concatenated and passed through a linear layer with GELU activation:
\begin{equation}
\mathbf{H}_{\text{integrated}} = \text{GELU}(\text{Linear}(\text{Concat}[\mathbf{H}'_1,\ldots,\mathbf{H}'_{N_j}])) \in \mathbb{R}^{B \times S \times D_i}.
\end{equation}

\subsubsection{BiGRU}
Temporal dependencies are captured using a stacked BiGRU with $L$ layers. The forward and backward passes yield
\begin{align}
\overrightarrow{\mathbf{H}}_{\text{GRU}} &= \text{GRU}_f(\mathbf{H}_{\text{integrated}}) \in \mathbb{R}^{B \times S \times D_g}, \\
\overleftarrow{\mathbf{H}}_{\text{GRU}} &= \text{GRU}_b(\mathbf{H}_{\text{integrated}}) \in \mathbb{R}^{B \times S \times D_g},
\end{align}
which are concatenated to form
\begin{equation}
\mathbf{H}_{\text{GRU}} = \text{Concat}[\overrightarrow{\mathbf{H}}_{\text{GRU}}, \overleftarrow{\mathbf{H}}_{\text{GRU}}] \in \mathbb{R}^{B \times S \times 2D_g}.
\end{equation}
This bidirectional design allows the model to capture both past and future context within each input window.

\subsubsection{Prediction}
The final representation from the last time step captures the activation state at the end of each input window:
\begin{equation}
\mathbf{h}_{\text{final}} = \mathbf{H}_{\text{GRU}}[:, -1, :] \in \mathbb{R}^{B \times 2D_g}.
\end{equation}
This is passed through a fully connected network to produce muscle activation predictions:
\begin{equation}
\hat{\mathbf{a}} = \text{FCN}(\mathbf{h}_{\text{final}}) \in \mathbb{R}^{B \times N_m}
\end{equation}

\subsection{Design of Loss Functions}
The proposed PI-MJCA-BiGRU employs a multi-objective loss function, $\mathcal{L}_{\text{total}}$, formulated as a weighted sum of three complementary components: a MSK dynamics loss ($\mathcal{L}_{\text{d}}$), a performance criterion loss ($\mathcal{L}_{\text{p}}$), and a boundary condition loss ($\mathcal{L}_{\text{b}}$). The total loss is expressed as:
\begin{equation}
\mathcal{L}_{\text{total}} = \omega_{\text{d}} \mathcal{L}_{\text{d}} + \omega_{\text{p}} \mathcal{L}_{\text{p}} + \omega_{\text{b}} \mathcal{L}_{\text{b}}
\label{eq:ltotal_enhanced}
\end{equation}
where $\omega_{\text{d}}$, $\omega_{\text{p}}$, and $\omega_{\text{b}}$ are hyperparameters that balance the relative contributions of each component. This formulation enables the network to simultaneously minimize muscle effort while maintaining dynamic consistency with observed movements and adhering to physiological constraints at each time step.

\subsubsection{MSK dynamics loss}
$\mathcal{L}_{\text{d}}$ reflects the underlying relationships among muscle activations and multi-joint kinematics in human motion under external forces, given by
\begin{equation}
\begin{aligned}
\mathcal{L}_{d} = \frac{1}{T} \sum_{t=1}^{T} \left\| \mathbf{M}(\mathbf{q}_t)\ddot{\mathbf{q}_t} + \mathbf{C}(\mathbf{q}_t, \dot{\mathbf{q}_t})\dot{\mathbf{q}_t} + \mathbf{G}(\mathbf{q}_t) - \boldsymbol{\tau}_{\text{e},t} - \boldsymbol{\tau}_{h,t} \right\|^{2},
\label{r1}
\end{aligned}
\end{equation}
where $\mathbf{q}_t \in \mathbb{R}^{N_j}$ represents the joint angles for all $N_j$ joints (assuming single degree of freedom per joint in the sagittal plane) at time step $t$. $\mathbf{M}(\mathbf{q}_t) \in \mathbb{R}^{N_j \times N_j}$ is the mass matrix, $\mathbf{C}(\mathbf{q}_t, \dot{\mathbf{q}}_t) \in \mathbb{R}^{N_j \times N_j}$ represents the Coriolis and centrifugal matrix, $\mathbf{G}(\mathbf{q}_t) \in \mathbb{R}^{N_j}$ is the gravity vector, and $\boldsymbol{\tau}_{e,t} \in \mathbb{R}^{N_j}$ denotes external torques. The joint velocities $\dot{\mathbf{q}}_t$ and accelerations $\ddot{\mathbf{q}}_t$ are obtained through discrete differentiation of the joint angles, and $T$ denotes the movement duration. The implementation details of the multi-joint dynamic system with external forces are provided in Section \ref{Biomechanics}.
$\boldsymbol{\tau}_{h,t} \in \mathbb{R}^{N_j}$ represents the joint torques generated by muscles, calculated as:
\begin{equation}
\boldsymbol{\tau}_{h,t} = \mathbf{R}_t^T \mathbf{F}_t^{mt}
\label{torque}
\end{equation}
where $\mathbf{R}_t \in \mathbb{R}^{N_m \times N_j}$ is the moment arm matrix with element $R_{n,j}$ representing the moment arm of the $n$-th muscle about the $j$-th joint exported from Opensim \cite{Seth-PLOSCB-2018}, $N_m$ is the total number of muscles, and $\mathbf{F}_t^{mt} \in \mathbb{R}^{N_m}$ is the muscle-tendon force vector.

The muscle-tendon force for the $n$-th muscle is calculated using the Hill-type muscle model with physiological parameters: maximum isometric force $F_{o,n}^{m}$, optimal fiber length $l_{o,n}^{m}$, optimal pennation angle $\varphi_{o,n}$, tendon slack length $l^{t}_{s,n}$, and maximum contraction velocity $v_{o,n}$, all initialized via scaling from generic OpenSim models. The muscle-tendon force is expressed as:
\begin{equation}
\begin{aligned}
F^{mt}_{t,n} &= (F_{t,n}^{CE} + F_{t,n}^{PE})\cos{\varphi_{t,n}}\\
&= F_{o,n}^{m}[\hat{a}_{t, n}f_{v}(\overline{v}_{t,n})f_{a}(\overline{l}^{m}_{t,n})+ f_{p}(\overline{l}^{m}_{t,n})]\cos{\varphi_{t,n}},
\label{Fmt}
\end{aligned}
\end{equation}
where $F_{t,n}^{CE}$ denotes the active contractile force generated by cross-bridge cycling, $F_{t,n}^{PE}$ represents the passive elastic force from stretched muscle fibers, and $\hat{a}_{t,n}$ is the neural activation predicted by the network. $f_{a}(\overline{l}^{m}_{t,n})$, $f_{v}(\overline{v}_{t,n})$, and $f_{p}(\overline{l}^{m}_{t,n})$ characterize the active force-length, force-velocity, and passive force-length relationships, respectively, with $\overline{l}^{m}_{t,n} = l^{m}_{t,n}/l_{o,n}^{m}$ and $\overline{v}_{t,n} = v_{t,n}/v_{o,n}$ being the normalized fiber length and velocity.
The pennation angle $\varphi_{t,n}$ evolves with muscle contraction according to:
\begin{equation}
\varphi_{t,n} = \sin^{-1}\left(\frac{l^m_{o,n}\sin{\varphi_{o,n}}}{l^{m}_{t,n}}\right).
\label{fi}
\end{equation}
While the muscle fiber length is updated by:
\begin{equation}
l^m_{t,n} = (l^{mt}_{t,n} - l^{t}_{t,n})/\cos{\varphi_{t,n}}
\label{lm}
\end{equation}
where $l^{t}_{t,n}$ is the tendon length, $l^{mt}_{t,n}$ represents the muscle-tendon length, approximated as polynomial functions of joint angles $q_t$ derived from OpenSim MSK geometry \cite{Seth-PLOSCB-2018}.

\subsubsection{Performance criterion loss}
$\mathcal{L}_p$ is formulated based on the performance criteria commonly used in MSK inverse dynamics optimization \cite{Schonhaut-TBME-2025}. By minimizing the sum of squared muscle activations, this loss function refines predictions to match observed movements while maintaining physiological feasibility:
\begin{equation}
\mathcal{L}_{p}= \frac{1}{T} \sum_{t=1}^{T}\sum_{n=1}^{N_m} (\hat{a}_{t, n})^2
\label{ploss}
\end{equation}
where $\hat{a}_{t, n}$ represents the predicted muscle activation of the $n$-th muscle at time step $t$.

\subsubsection{Boundary condition loss}
$\mathcal{L}_b$ ensures muscle activation predictions remain within physiologically plausible bounds $[0.01, 1]$ by penalizing boundary violations:
\begin{equation}
\mathcal{L}_b = \frac{1}{T} \sum_{t=1}^{T} \sum_{n=1}^{N_m} \left[ \max(0, 0.01 - \hat{a}_{t,n})^2 + \max(0, \hat{a}_{t,n} - 1)^2 \right]
\label{boundary}
\end{equation}
where the first term penalizes activations below the minimum threshold and the second term penalizes those exceeding the maximum bound, ensuring biologically realistic predictions.

\subsection{Multi-joint dynamic system}\label{Biomechanics}
\begin{figure}[htbp]
    \centering
    \includegraphics[width=0.3\textwidth]{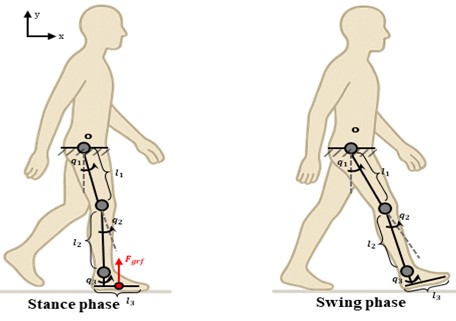}
    \caption{Planar model of the lower limb during the stance and swing phases of gait with 3 joints (Hip, knee, ankle).}
    \label{Gait}
\end{figure}

This study employs a planar 3-DOF lower-limb model in which the hip joint is designated as the fulcrum, serving as the origin of the global reference frame $\{W\}$ (x-axis: forward, y-axis: upward). 
The lower limb is modeled as a planar kinematic chain with joint index set $\mathcal{J}=\{1,2,3\}$ corresponding to hip, knee, and ankle, respectively. 
The system configuration is described by the joint vector $\mathbf{q}=[q_1,q_2,q_3]^T$, where $q_1$, $q_2$, and $q_3$ are the generalized coordinates representing the hip, knee, and ankle joint angles in the sagittal plane, with positive values defined as counterclockwise rotations.
Segments are indexed by $\mathcal{K}=\{1,2,3\}$, representing thigh, shank, and foot. 
Each segment $k\in\mathcal{K}$ is characterized by its mass $m_k$, moment of inertia about the center of mass $I_k$, length, 
and a center-of-mass position vector $\mathbf{r}_{\mathrm{com},k}$ expressed in the local coordinate frame of its proximal joint. 

The model is based on the following assumptions: (i) each segment behaves as a rigid body; (ii) joints are ideal revolutes; (iii) the foot is treated as rigid, and deformation is neglected; and (iv) no slip occurs at the foot–ground interface during stance.  
The dynamic parameters of each limb are obtained by scaling the widely used 2392-OpenSim model in accordance with the subject’s height and weight. Unless otherwise specified, joint states and all associated dynamic terms (e.g., inertia, Coriolis, gravity, and torques) are expressed in the global frame $\{W\}$.

The equation of the lower limb model could be given by:
\begin{equation}
\mathbf{M}(\mathbf{q}) \ddot{\mathbf{q}} + \mathbf{C}(\mathbf{q}, \dot{\mathbf{q}})\dot{\mathbf{q}} + \mathbf{G}(\mathbf{q}) 
= \boldsymbol{\tau}_{h} + \boldsymbol{\tau}_{grf},
\end{equation}
where $\mathbf{M}(\mathbf{q})$ is the inertia matrix, $\mathbf{C}(\mathbf{q},\dot{\mathbf{q}})$ captures Coriolis and centrifugal effects, and $\mathbf{G}(\mathbf{q})$ is the gravitational torque vector. 
On the right-hand side, $\boldsymbol{\tau}_h = [\tau_{h1},\,\tau_{h2},\,\tau_{h3}]^T$ represents joint torques generated by muscle–tendon units, while $\boldsymbol{\tau}_{grf}$ denotes torques induced by ground reaction forces (GRF). 
Tangential friction torques are neglected, consistent with the no-slip assumption.

The gait cycle dynamics depend on the contact condition:
\begin{equation}
\boldsymbol{\tau}_{h} =
\begin{cases}
\mathbf{M}(\mathbf{q}) \ddot{\mathbf{q}} + \mathbf{C}(\mathbf{q}, \dot{\mathbf{q}})\dot{\mathbf{q}} + \mathbf{G}(\mathbf{q}), & \text{swing} \\[0.5em]
\mathbf{M}(\mathbf{q}) \ddot{\mathbf{q}} + \mathbf{C}(\mathbf{q}, \dot{\mathbf{q}})\dot{\mathbf{q}} + \mathbf{G}(\mathbf{q}) - \boldsymbol{\tau}_{grf}, & \text{stance}.
\end{cases}
\end{equation}

During the swing phase, joint torques are governed solely by inertial and gravitational effects. 
The inertia matrix $\mathbf{M}(\mathbf{q}) \in \mathbb{R}^{3\times 3}$ is symmetric and its elements $M_{ij}$ (with $i,j \in \mathcal{J}$, representing inertial coupling between the $i$-th and $j$-th joints) are derived from the Lagrangian formulation based on the total kinetic energy:
\begin{equation}
M_{ij} = \frac{\partial^2 E}{\partial \dot{q}_i \partial \dot{q}_j},
\end{equation}
where $E$ is the sum of translational and rotational kinetic energy of all segments:
\begin{equation}
E = 
\sum_{k=1}^{3} \left( \frac{1}{2} m_k (\dot{x}_{com,k}^2 + \dot{y}_{com,k}^2) + \frac{1}{2} I_k \dot{\theta}_k^2 \right).
\end{equation}
Here $(\dot{x}_{com,k}, \dot{y}_{com,k})$ is the translational velocity of the center of mass of segment $k$, computed from the kinematic chain, and $\dot{\theta}_k$ is the angular velocity of segment $k$, equal to the cumulative sum of proximal joint velocities:
\begin{equation}
\dot{\theta}_k = \sum_{i=1}^{k} \dot{q}_i.
\end{equation}

The Coriolis matrix $\mathbf{C}(\mathbf{q}, \dot{\mathbf{q}}) \in \mathbb{R}^{3\times 3}$ captures velocity-dependent coupling effects between joints. Its elements $C_{ij}$ (with $i,j \in \mathcal{J}$) are computed using the Christoffel symbols of the first kind:
\begin{equation}
C_{ij} = \sum_{r=1}^{3} \Gamma_{ij}^r \dot{q}_r, 
\quad i,j,r \in \mathcal{J},
\end{equation}
where
\begin{equation}
\Gamma_{ij}^r = \tfrac{1}{2} \left(
\frac{\partial M_{ij}}{\partial q_r} +
\frac{\partial M_{ir}}{\partial q_j} -
\frac{\partial M_{jr}}{\partial q_i}
\right), 
\quad i,j,r \in \mathcal{J}.
\end{equation}

The gravitational torque vector $\mathbf{G}(\mathbf{q}) \in \mathbb{R}^{3}$ 
accounts for the effect of gravity on each joint. 
Its elements $G_i$ (with $i \in \mathcal{J}$) are obtained as the gradient of the potential energy:
\begin{equation}
G_i = \frac{\partial}{\partial q_i} 
\sum_{k=1}^{3} m_k g \, y_{\mathrm{com},k}, 
\quad i \in \mathcal{J}, \; k \in \mathcal{K},
\end{equation}
where $y_{com,k}$ is the vertical position of the $k$-th segment’s center of mass.

During stance, the GRF is applied at the Center of Pressure (COP) under the rigid-foot assumption.  
While joint states and dynamic terms are expressed in the global frame $\{W\}$ as established earlier, the GRF and COP are initially measured in the force plate frame $\{P\}$ and thus require explicit transformation into $\{W\}$.

The COP position in the world frame is obtained as
\begin{equation}
\mathbf{p}_{cop}^{W} = 
\begin{bmatrix}
x_{cop}^W & y_{cop}^W
\end{bmatrix}^T
= {}^{W}\!T_{P}\,\mathbf{p}_{cop}^{P},
\end{equation}
where $\mathbf{p}_{cop}^{P}$ is the measured COP in $\{P\}$, and ${}^{W}\!T_{P}$ is the calibrated rigid-body transformation from $\{P\}$ to $\{W\}$.  
The resulting joint torques induced by the GRF are obtained via the principle of virtual work:
\begin{equation}
\boldsymbol{\tau}_{grf} = \mathbf{J}_v(\mathbf{q})^T \,\mathbf{F}_{grf},
\end{equation}
where $\boldsymbol{\tau}_{grf} \in \mathbb{R}^3$ is the torque vector applied at the joints due to the GRF, $\mathbf{F}_{grf} = \begin{bmatrix} F_x^W & F_y^W \end{bmatrix}^T \in \mathbb{R}^2$ is the ground reaction force expressed in $\{W\}$, and $\mathbf{J}_v(\mathbf{q}) \in \mathbb{R}^{2\times 3}$ is the Jacobian that maps joint velocities $\dot{\mathbf{q}}$ to the Cartesian velocity of the COP in $\{W\}$.

\section{Dataset and Experimental Settings}\label{Dataset}
\subsection{Benchmark Dataset}
The publicly available biomechanics dataset from Camargo et al. \cite{CAMARGO2021110320} was utilized in this study. This comprehensive dataset contains biomechanical and wearable sensor data from adults performing multiple locomotion modes, including treadmill walking at different speeds ranging from 0.5 to 1.85 m/s. From this dataset, treadmill walking data from six subjects (4 males, 2 females; age: $20.8 \pm 1.0$ years; height: $1.69 \pm 0.09$ m; body mass: $67.7 \pm 10.8$ kg) were selected.

The original data were collected using a VICON motion capture system at 200 Hz with the Helen Hayes Hospital marker set, synchronized with Bertec force plates at 1000 Hz. For our analysis, data at three specific walking speeds (0.9 m/s, 1.3 m/s, and 1.7 m/s) were extracted. The raw data were processed using the STRIDES script provided by the authors to segment continuous treadmill walking trials into individual gait cycles based on right heel strike events. During segmentation, only steady-state walking segments were retained, excluding acceleration and deceleration periods.

For MSK analysis, the OpenSim gait2392 model was scaled to match each subject's anthropometry. Joint kinematics were computed using the Inverse Kinematics tool.
SO was then applied to estimate muscle forces and activations, providing the biomechanically consistent reference solutions against which our PI-MJCA-BiGRU predictions are compared.

\subsection{Self-Collected Dataset}
\begin{figure}[htbp]
    \centering
    \includegraphics[width=0.38\textwidth]{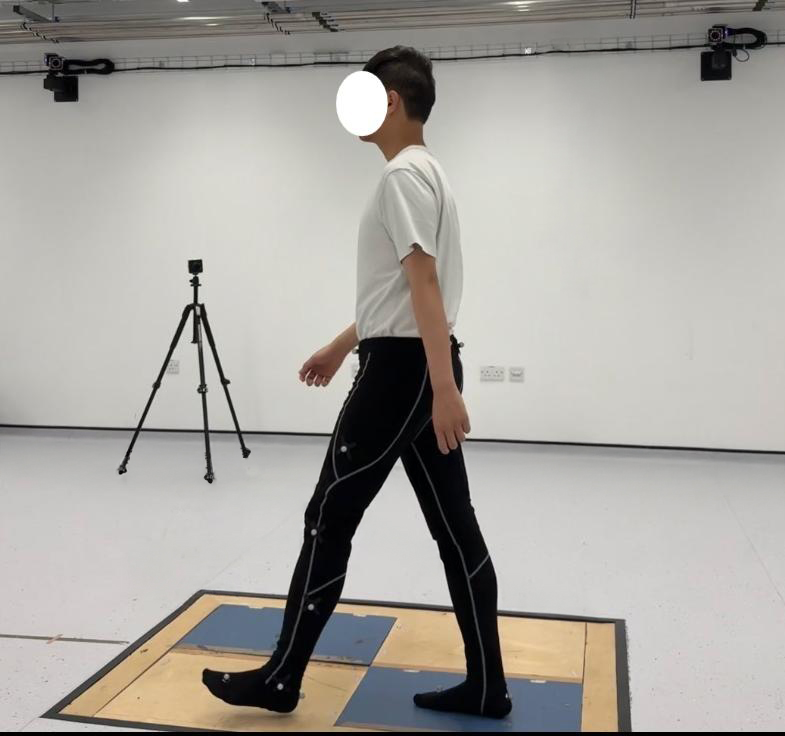}
    \caption{Subject walking on an instrumented platform with reflective markers for motion capture and ground reaction force measurement.}
    \label{experiments}
\end{figure}

\begin{figure*}[!htbp]
  \centering
  \includegraphics[width=0.95\textwidth]{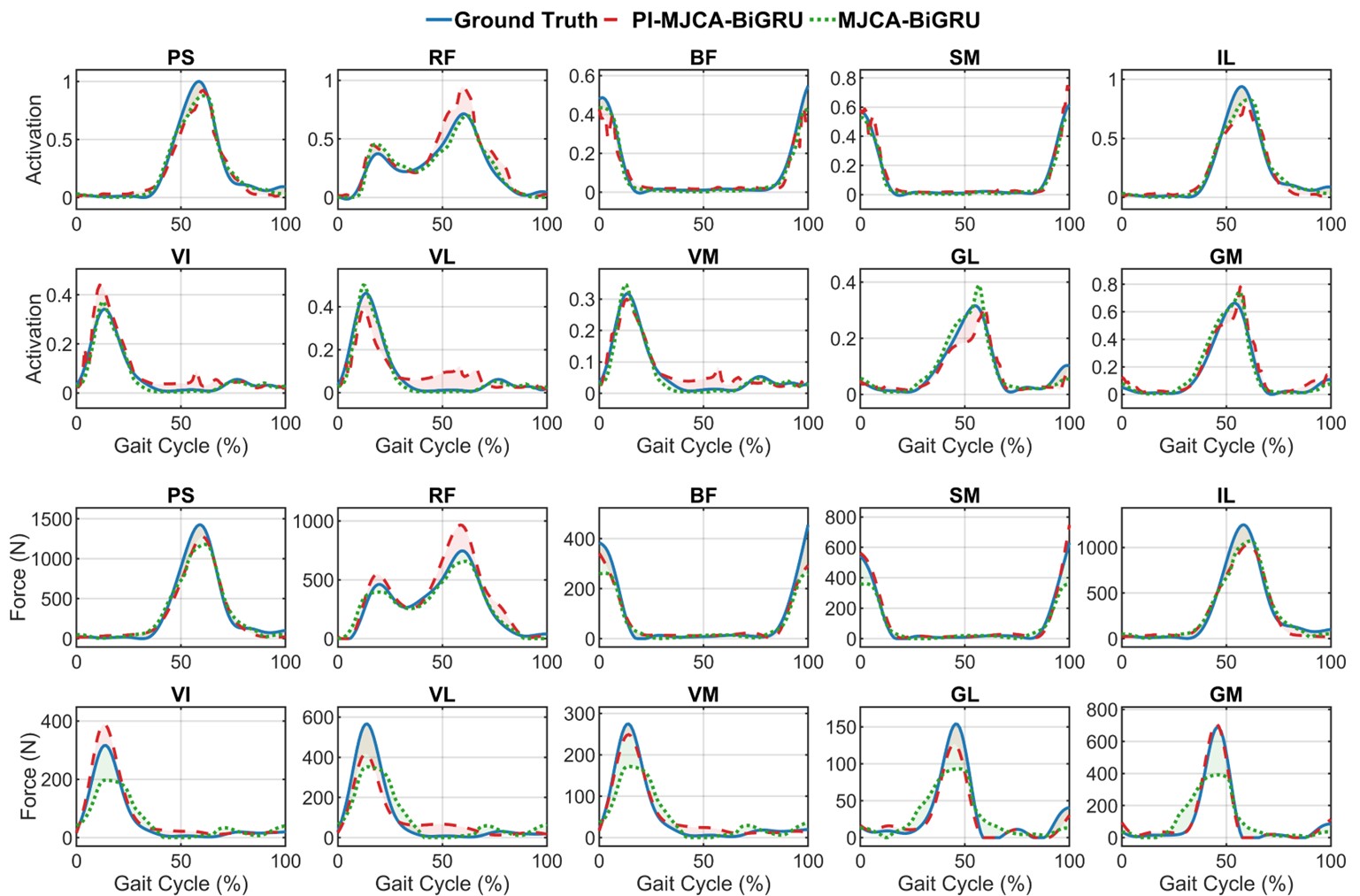} 
  \caption{Representative results of the gait cycle case through our method. The predicted outputs include the muscle activations and muscle forces of the ten muscles. The results show that PI-MJCA-BiGRU achieves comparable tracking accuracy to supervised MJCA-BiGRU, closely matching the ground truth.}
  \label{fig:representative}
\end{figure*}
The experiment was approved by the Engineering and Physical Sciences Faculty Research Ethics Committee of the University of Leeds (LTELEC-001). Six subjects (5 males, 1 female; age: $26.3 \pm 1.0$ years; height: $1.75 \pm 0.07$ m; body mass: $72.7 \pm 11.0$ kg) participated after providing written informed consent.
Motion capture was performed using a VICON system (Oxford Metrics, UK) synchronized with an AMTI force platform (Advanced Mechanical Technology Inc., USA), both sampling at 250 Hz. Sixteen reflective markers were placed according to the VICON Plug-in-Gait lower body protocol. The experimental setup is illustrated in Fig. \ref{experiments}.
Each subject performed 10 successful trials, where a successful trial was defined as one in which the participant's foot landed completely within the boundaries of the force platform without targeting or altering their gait pattern.
For each trial, we extracted one complete gait cycle of the leg that contacted the force platform, defined from initial contact (heel strike) to the subsequent initial contact of the same foot. 

Data processing involved marker labeling and gap-filling in VICON Nexus, followed by the same MSK analysis pipeline as introduced in the Benchmark Dataset.

\subsection{Dataset Preparation and Training Protocol}

\subsubsection{Data Processing Pipeline}
From both datasets, we extracted data corresponding to the right leg for our planar 3-DOF model. Joint kinematics (velocities and accelerations) for the hip, knee, and ankle were computed through finite-difference differentiation of the measured angle trajectories. 
From the full-body 2392 MSK model, we selected ten muscles critical for single-leg gait analysis: psoas major (PS), rectus femoris (RF), biceps femoris long head (BF), semimembranosus (SM), iliacus (IL), vastus intermedius (VI), vastus lateralis (VL), vastus medialis (VM), gastrocnemius lateralis (GL), and gastrocnemius medialis (GM). The corresponding muscle activations and forces from SO were extracted for these muscles, forming the input-output pairs for model evaluation.
To prepare sequential inputs for the recurrent architecture, we applied a sliding-window strategy with a window size of 20 and a stride of 2, ensuring overlap while preserving sequential continuity.

\subsubsection{Training Configuration}
The MJCA-BiGRU was configured with joint feature dimension $D_l$ = 64, attention heads $N_h$ = 2, integrated dimension $D_i$ = 128, and BiGRU hidden size $D_g$ = 128 (256 bidirectional). The network used $L$ = 2 BiGRU layers (dropout 0.1) and output layers with dimensions 256, 128, 64, and 10.

All models were trained for 500 epochs with a batch size of 8 using the Adam optimizer with an initial learning rate $\eta$ = 5 $\times$ 10$^{-4}$, weight decay of 0.01, and a cosine annealing schedule decreasing to 10\% of the initial rate. 
The physics-informed loss weights were empirically determined through grid search: $\omega_d$ = 3 for dynamics constraints, $\omega_p$ = 1000 for activation penalty, and $\omega_b$ = 500 for physiological boundaries. For model evaluation, unless otherwise specified, an intra-condition split of 80\% for training and 20\% for testing was applied to all experiments across both datasets.

\subsection{Baseline Methods}
\begin{figure*}[!htbp]
  \centering
  \includegraphics[width=\textwidth]{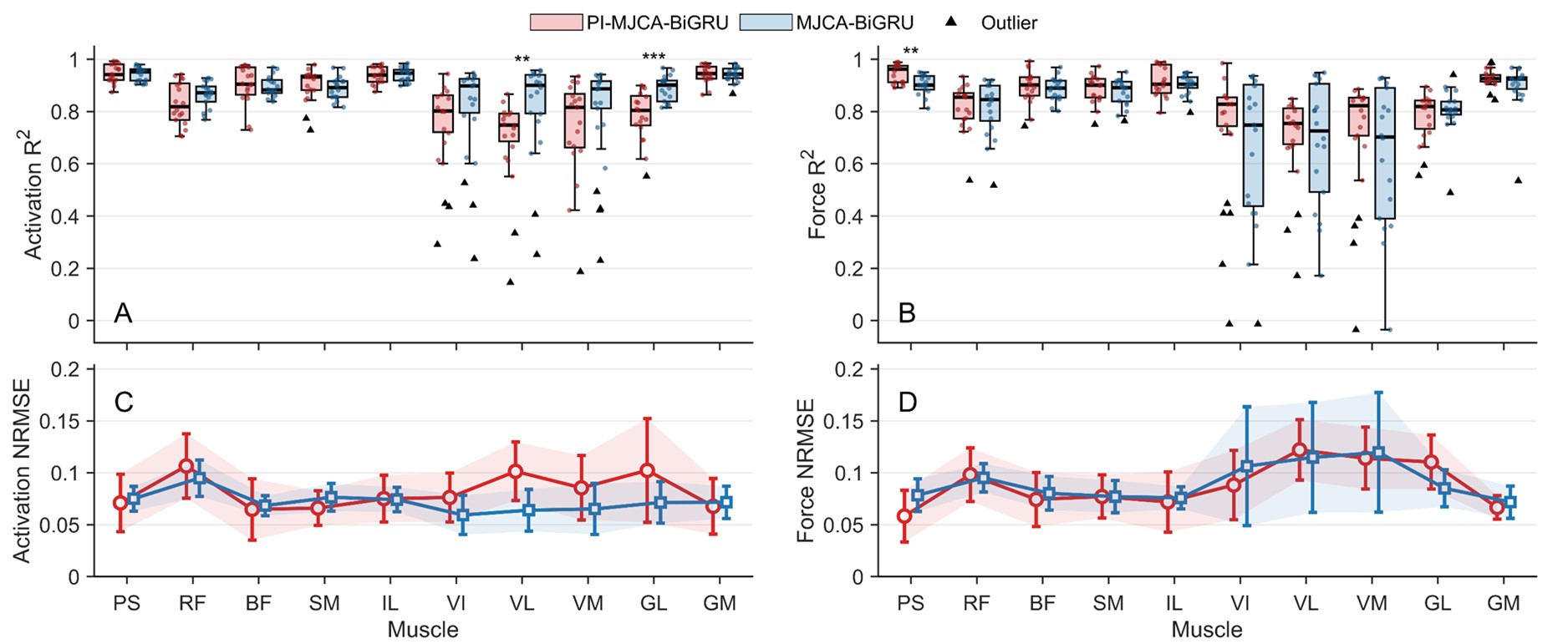}
  \caption{Performance comparison of physics-informed (PI-MJCA-BiGRU) and supervised (MJCA-BiGRU) training approaches for muscle activations and forces prediction on the \textbf{benchmark dataset}. Results are presented for activation R$^2$ (A), force R$^2$ (B), activation NRMSE (C), and force NRMSE (D) across ten muscles. "$\blacktriangle$" indicates outliers in the distribution. Overall, PI-MJCA-BiGRU achieves accuracy comparable to supervised MJCA-BiGRU training.}
  \label{fig:overall_pd}
\end{figure*}

\subsubsection{Supervised Learning Baseline}
To evaluate our physics-informed training paradigm, we compared it against a supervised baseline using an identical MJCA-BiGRU architecture trained with mean squared error (MSE) loss on SO-derived muscle activations and forces from OpenSim.

\subsubsection{Alternative Architecture Configurations for Ablation Study}
\label{Alternative Architecture Configurations}
%
%
%
To validate the effectiveness of the MJCA module, we conducted ablation studies with two alternative configurations. The first (Pi-BiGRU-only) removes the MJCA module and directly feeds concatenated joint features into the BiGRU layers. 
The second (PI-CNN-BiGRU) replaces the MJCA module with three consecutive convolutional layers (64, 128, and 128 channels respectively, with kernel size 3, stride 1, padding 1), each followed by GELU activation and dropout (0.05), before passing the output to the same BiGRU backbone and output layers.
Both baseline architectures used identical BiGRU specifications, output layers, and training hyperparameters, and were trained with the same physics-informed loss function.

\subsection{Evaluation Criteria}
Three metrics were employed to evaluate model performance: coefficient of determination (R$^2$) and normalized root mean square error (NRMSE).

The coefficient of determination measures the proportion of variance explained by the model:
\begin{equation}
R^2 = 1 - \frac{\sum_{m=1}^{M} (Y_m - \hat{Y}_m)^2}{\sum_{m=1}^{M} (Y_m - \bar{Y}_m)^2}
\end{equation}
where $M$ is the total number of samples, $Y_m$ and $\hat{Y}_m$ are actual and predicted values, and $\bar{Y}_m$ is the mean of actual values. R$^2$ approaches 1 for accurate predictions.

NRMSE normalizes the prediction error by the data range:
\begin{equation}
NRMSE = \frac{\sqrt{\frac{1}{M} \sum_{m=1}^{M} (Y_{m} - \hat{Y}_{m})^2}}{Y_{max} - Y_{min}}
\end{equation}
where $Y_{max}$ and $Y_{min}$ are the maximum and minimum actual values. Lower NRMSE indicates better performance.

\section{Results}\label{results}
\begin{figure*}[!htbp]
  \centering
  \includegraphics[width=\textwidth]{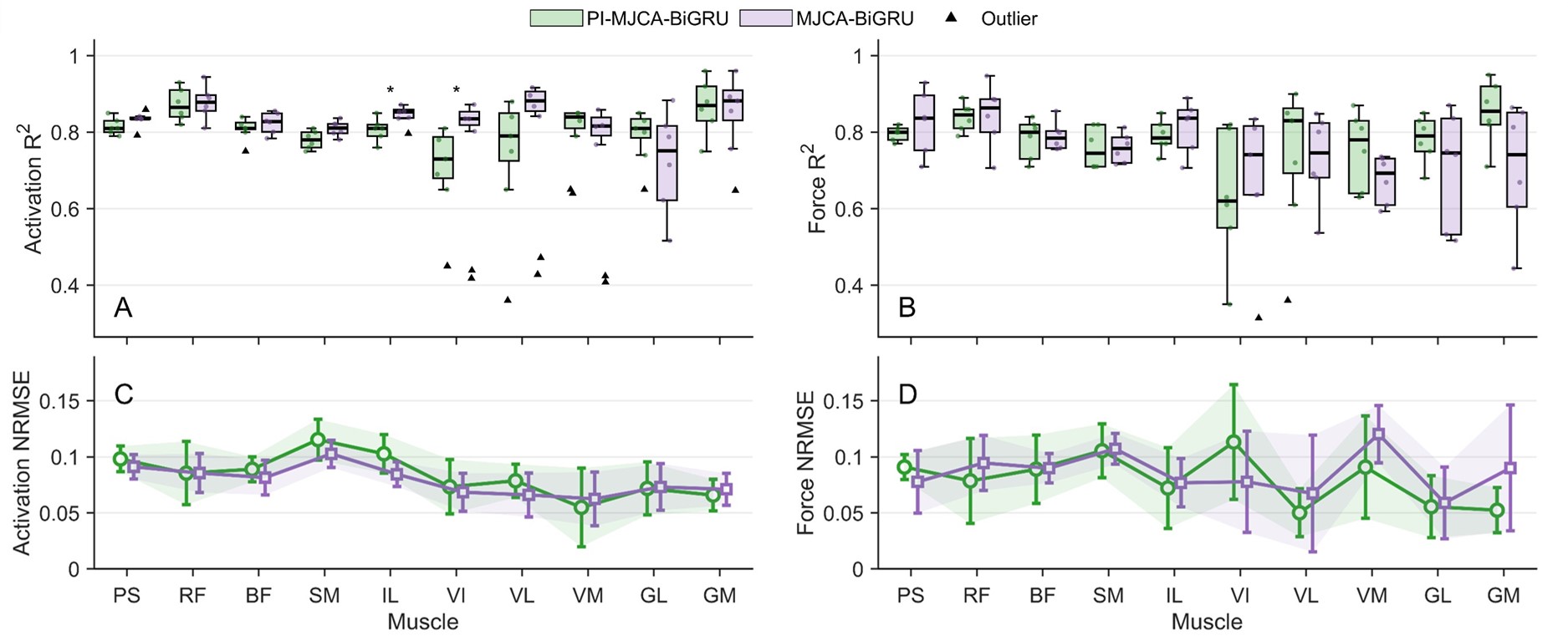} 
  \caption{Performance comparison on the \textbf{self-collected dataset}. Results show activation R$^2$ (A), force R$^2$ (B), activation NRMSE (C), and force NRMSE (D) across ten muscles. Overall, PI-MJCA-BiGRU achieves accuracy comparable to supervised MJCA-BiGRU training.}
  \label{fig:overall_sd}
\end{figure*}

\begin{figure*}[!htbp]
  \centering
  \includegraphics[width=\textwidth]{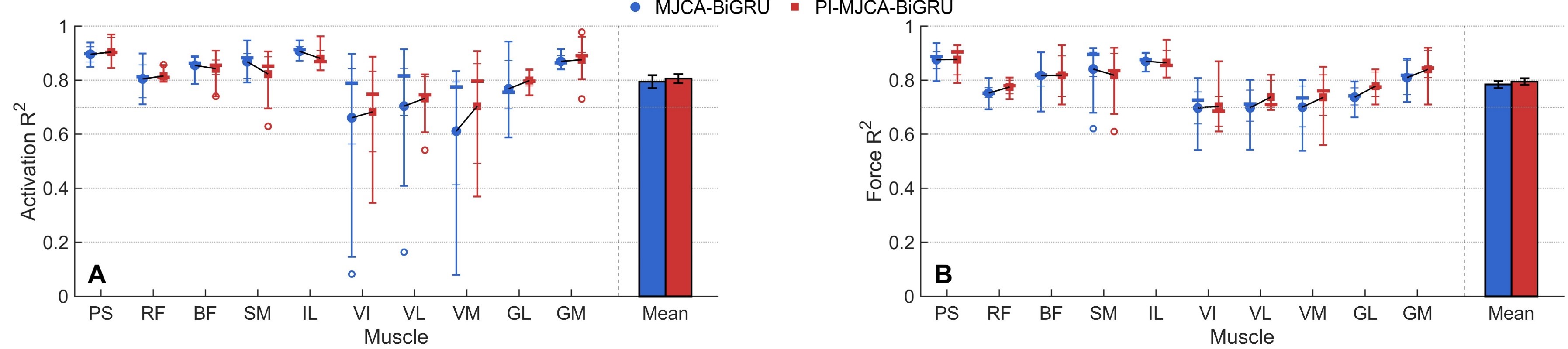} 
  \caption{LOSO cross-validation results evaluating generalization performance on the \textbf{benchmark dataset}. Results show activation R$^2$ (A) and force R$^2$ (B) across ten muscles.}
  \label{fig:LOSO}
\end{figure*}

In this section, we evaluate the performance of our proposed PI-MJCA-BiGRU model trained with physics-informed loss functions using both the benchmark dataset and our self-collected dataset. The experimental results are organized into two main components. Section \ref{sec:supervised_comparison} presents a comprehensive comparison between our physics-informed training approach and supervised learning methods, where both utilize identical MJCA-BiGRU architecture but differ in their training paradigms—physics-informed loss ($\mathcal{L}_{total}$) versus data-driven MSE loss. 
This comparison includes overall performance metrics across both datasets, Leave-One-Speed-Out (LOSO) cross-validation to assess generalization capability, and feature visualization analysis to examine the learned representations. Section \ref{sec:ablation} provides ablation studies investigating two critical aspects: the contribution of individual loss components to the overall performance, and the effectiveness of the MJCA module. All experiments were implemented using PyTorch and executed on an NVIDIA RTX 3070ti GPU with 16 GB memory.

\subsection{Comparison with Supervised Learning Methods}\label{sec:supervised_comparison}

\subsubsection{Overall Performance Comparison}
Fig. \ref{fig:representative} demonstrates that PI-MJCA-BiGRU (dashed red) achieves comparable tracking accuracy to supervised MJCA-BiGRU (dotted green) against ground truth (solid blue) throughout the gait cycle. Both methods accurately capture muscle-specific activation patterns and peak timings, with force predictions showing similar fidelity.

To quantitatively assess this comparable performance, we evaluated both methods using R$^2$ and NRMSE metrics across two datasets. On the benchmark dataset (Fig. \ref{fig:overall_pd}), PI-MJCA-BiGRU achieved mean activation R$^2$ values exceeding 0.8 for most muscles, matching the supervised approach's performance. Force prediction R$^2$ values similarly ranged from 0.7 to 0.95 for both methods. The NRMSE values remained consistently below 0.15 across all muscles, confirming low prediction errors for both training paradigms. While statistical differences were observed in specific muscles (VL: $p < 0.01$, GL: $p < 0.001$), the overall performance demonstrated that physics-informed training achieves accuracy comparable to supervised learning.

The self-collected dataset (Fig. \ref{fig:overall_sd}) exhibited similar performance patterns, though with generally lower R$^2$ values reflecting the increased variability of overground walking. PI-MJCA-BiGRU maintained robust performance with mean activation R$^2$ values ranging from 0.75 to 0.9, comparable to the supervised approach across all muscles.  Force predictions showed greater inter-muscle variability, particularly in muscles (GL, GM), yet both methods maintained NRMSE values below 0.15. These consistent results across different walking conditions confirm the robustness of physics-informed training.

These comprehensive evaluations demonstrate that physics-informed training achieves performance comparable to supervised learning across multiple metrics and datasets, validating that biomechanical constraints can effectively guide network training without requiring computationally expensive ground truth labels from computational MSK models.

\begin{figure*}[!htbp]
  \centering
  \includegraphics[width=\textwidth]{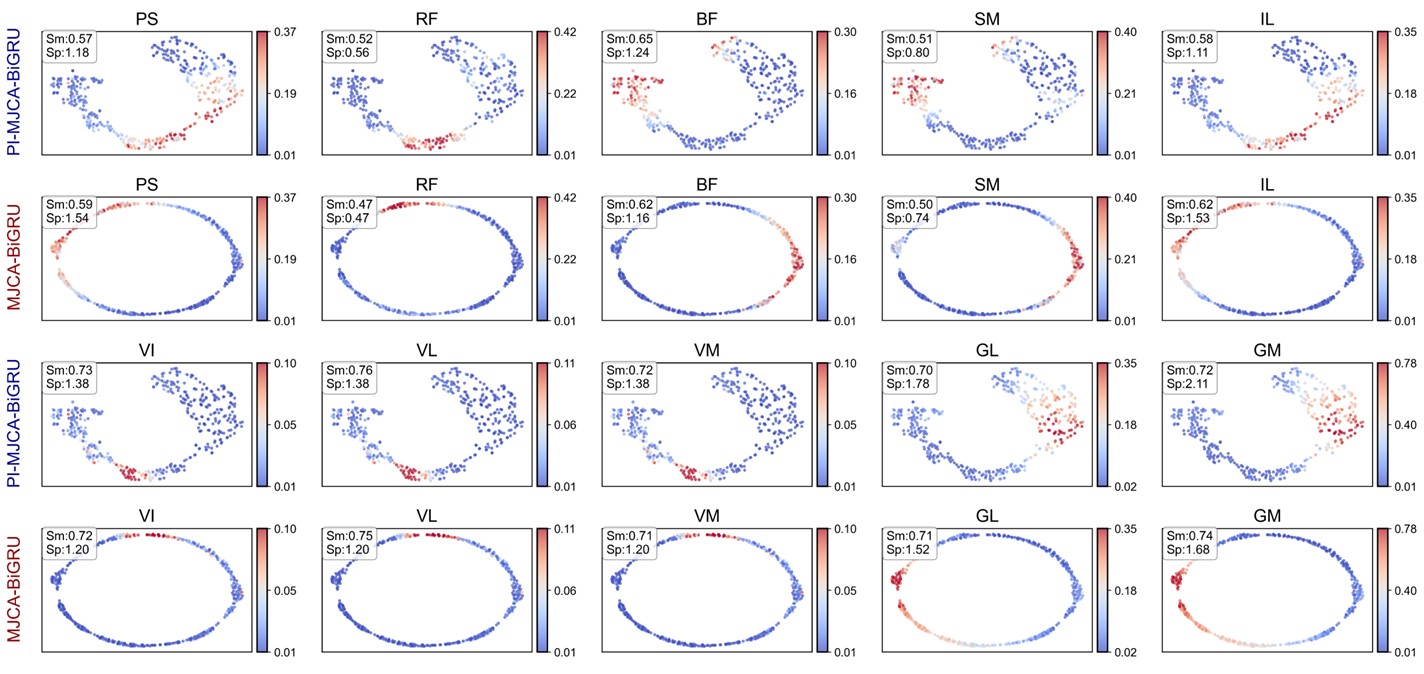} 
  \caption{t-SNE visualization of learned feature representations from a representative trial, comparing PI-MJCA-BiGRU and supervised MJCA-BiGRU across ten muscles. Each point represents a test sample, colored by muscle activation magnitude. Procrustes similarity = 0.889, demonstrating comparable representation quality between training paradigms.}
  \label{fig:muscle_comparison_grid}
\end{figure*}

\subsubsection{LOSO Cross-Validation}
To further evaluate generalization capability, we conducted LOSO cross-validation on the benchmark dataset. Models trained on 0.9 m/s and 1.7 m/s data were tested on the intermediate 1.3 m/s speed, with all training configurations held constant.

Relative to single-speed evaluation (Fig.~\ref{fig:overall_pd}, mean R$^{2}>$0.85), 
LOSO reduced performance to 0.8 for both activation and force prediction 
(Fig.~\ref{fig:LOSO}), as expected for cross-speed generalization. 
Mean performance was comparable across methods, but muscle-level distributions differed: 
PI-MJCA-BiGRU showed tighter confidence intervals and fewer outliers across most muscles, 
whereas MJCA-BiGRU exhibited larger variance. 
The black connecting lines indicate higher R$^{2}$ for PI-MJCA-BiGRU on most muscles; gains are 
especially visible for activation in VI, VL, VM, and GL. 
Overall, the PI constraints improve across-muscle consistency and robustness rather than shifting the mean alone.
These results indicate that physics-informed constraints yield more robust representations by capturing speed-invariant biomechanical principles, thereby reducing overfitting to speed-specific patterns and improving generalization to unseen walking speeds.

\subsubsection{Feature Visualization}
To evaluate PI-MJCA-BiGRU's feature learning capabilities without labels, we performed t-SNE visualization of learned representations across all subjects and computed three metrics: Procrustes similarity for geometric correspondence, local smoothness (Sm), and separation index (Sp). Results from a representative trial are presented below, with similar patterns observed consistently across all tested conditions.

Fig.~\ref{fig:muscle_comparison_grid} and Table~\ref{tab:feature_metrics} reveal that despite lacking labels, PI-MJCA-BiGRU achieved high Procrustes similarity (0.889) with traditional supervised learning, indicating effective physics-guided feature discovery. Fig.~\ref{fig:muscle_comparison_grid} shows distinct geometric patterns: PI-MJCA-BiGRU produces C-shaped manifolds reflecting gait cyclicity, while supervised MJCA-BiGRU generates circular patterns characteristic of data-driven optimization. Table~\ref{tab:feature_metrics} quantifies these differences. Both methods achieved comparable smoothness (0.65 vs 0.64), but PI-MJCA-BiGRU demonstrated superior separation for ankle flexors (GL: 1.78 vs 1.52; GM: 2.11 vs 1.68). Conversely, both approaches showed lower separation for bi-articular muscles (RF: 0.56 vs 0.47; SM: 0.80 vs 0.74), reflecting the inherent complexity of muscles spanning multiple joints.

The high feature similarity with comparable separation metrics validates PI-MJCA-BiGRU as an effective alternative when labeled data is unavailable, while offering additional interpretability through physically grounded representations.

\begin{table}[htbp]
\centering
\footnotesize 
\caption{Feature space metrics comparing PI-MJCA-BiGRU and MJCA-BiGRU across ten muscles from a representative trial}
\label{tab:feature_metrics}
\begin{tabular}{lcccc}
\toprule
\multirow{2}{*}{Muscle} & \multicolumn{2}{c}{Sm} & \multicolumn{2}{c}{Sp} \\
\cmidrule(lr){2-3} \cmidrule(lr){4-5}
& PI & S & PI & S \\
\midrule
PS & $0.57$ & $0.59$ & $1.18$ & $1.54$ \\
RF & $0.52$ & $0.47$ & $0.56$ & $0.47$ \\
BF & $0.65$ & $0.62$ & $1.24$ & $1.16$ \\
SM & $0.51$ & $0.50$ & $0.80$ & $0.74$ \\
IL & $0.58$ & $0.62$ & $1.11$ & $1.53$ \\
VI & $0.73$ & $0.72$ & $1.38$ & $1.20$ \\
VL & $0.76$ & $0.75$ & $1.38$ & $1.20$ \\
VM & $0.72$ & $0.71$ & $1.38$ & $1.20$ \\
GL & $0.70$ & $0.71$ & $1.78$ & $1.52$ \\
GM & $0.72$ & $0.74$ & $2.11$ & $1.68$ \\
\midrule
\textbf{Mean} & \textbf{0.65$\pm$0.09} & 0.64$\pm$0.10 & \textbf{1.29$\pm$0.42} & 1.22$\pm$0.35 \\
\bottomrule
\end{tabular}
\begin{tablenotes}
\footnotesize 
\item Note: PI: PI-MJCA-BiGRU, S: MJCA-BiGRU.
\end{tablenotes}
\end{table}



\subsection{Ablation Studies}\label{sec:ablation}

\subsubsection{Loss Component Analysis}
To investigate the contribution of each loss component, we conducted ablation experiments with consistent architecture and training strategies, systematically excluding individual terms from $\mathcal{L}_{\text{total}}$. Table \ref{tab:loss-ablation} shows the mean R$^2$ and NRMSE across all subject-condition-muscle combinations from the benchmark test set.

The complete model with all three loss components achieved optimal performance. 
The exclusion of the MSK dynamics loss $\mathcal{L}_d$ resulted in severe performance degradation, with R$^2$ values decreasing to $0.11\pm0.56$ (activation) and $0.09\pm0.44$ (force), demonstrating its indispensable role in embedding biomechanical constraints into the learning process.
The removal of the performance criterion loss $\mathcal{L}_p$ led to substantial deterioration with R$^2$ values decreasing to $0.65\pm0.33$ (activation) and $0.63\pm0.16$ (force), demonstrating its critical role in guiding the network toward physiologically optimal muscle recruitment patterns.
The absence of the boundary condition loss $\mathcal{L}_b$ caused a modest decline, suggesting its role in maintaining
activation bounds [0.01, 1] for biological plausibility. 
These findings establish a clear hierarchical importance among loss components ($\mathcal{L}_d \gg \mathcal{L}_p > \mathcal{L}_b$).


\subsubsection{Network Architecture Analysis}
To isolate the contribution of our proposed MJCA module, we conducted ablation studies comparing PI-MJCA-BiGRU against two alternative architectures. The baseline configurations, detailed in Section.\ref{Alternative Architecture Configurations}, include: PI-BiGRU-only and PI-CNN-BiGRU. All three models employ identical BiGRU specifications, output projections, and physics-informed loss functions.

\begin{figure}[htbp]
    \centering
    \includegraphics[width=0.5\textwidth]{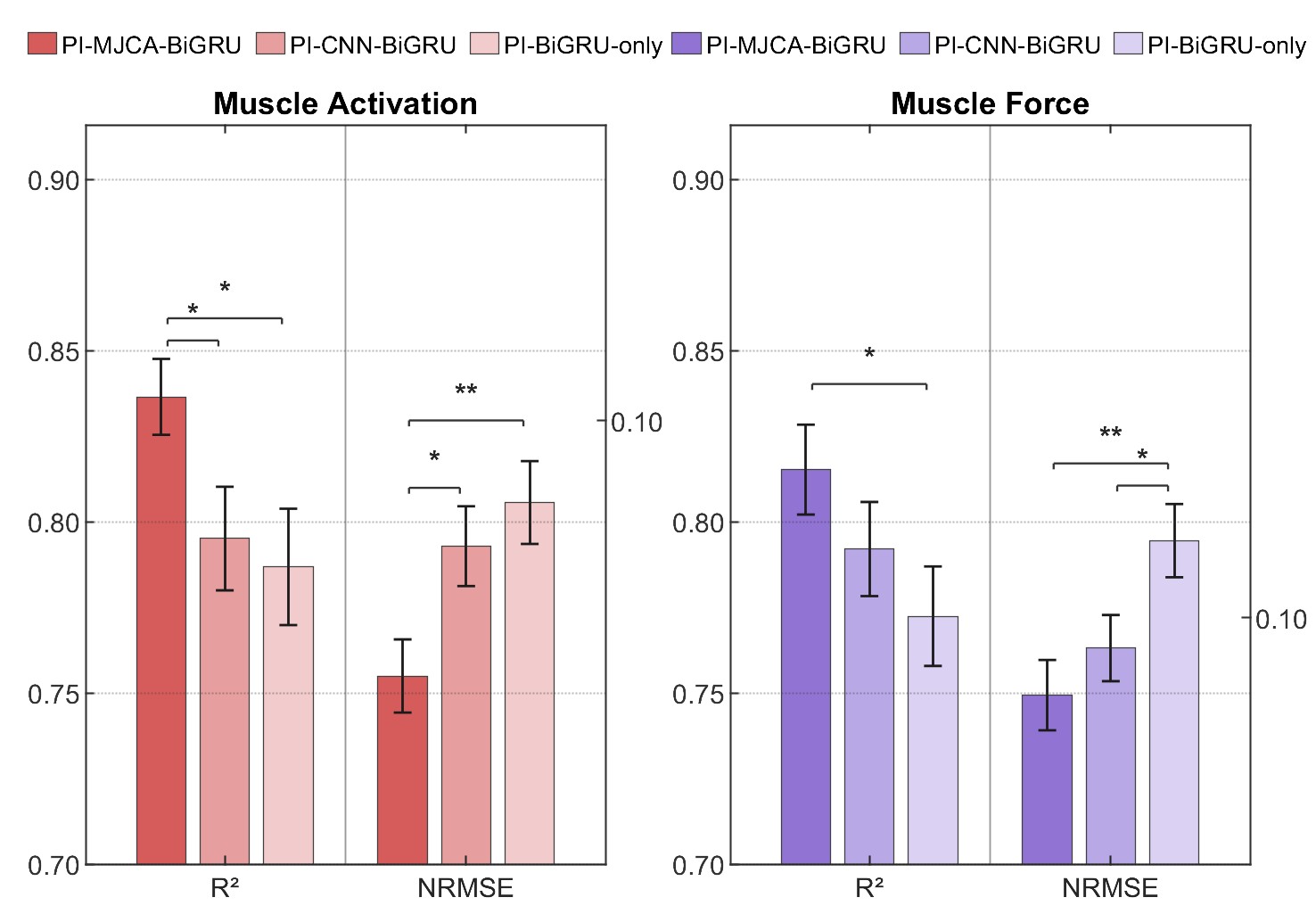}
    \caption{Comparative performance of deep learning architectures for muscle activations and forces estimation on the benchmark dataset. 
    ($^{*}p < 0.05$, $^{**}p < 0.01$;)}
    \label{fig:ablation}
\end{figure}

Fig.~\ref{fig:ablation} presents the ablation study results comparing three physics-informed architectures trained and evaluated on the benchmark dataset across all subjects, walking speeds, and muscles. As shown in the left panel, PI-MJCA-BiGRU achieved superior activation prediction performance with a mean R$^2$ = 0.84, representing 5.5\% and 6.2\% improvements over PI-CNN-BiGRU and PI-BiGRU-only, respectively. The corresponding NRMSE reductions further confirm the advantage of attention-based modeling.
For force prediction (right panel), all models showed lower R$^2$ values compared to activation prediction. 
Nevertheless, PI-MJCA-BiGRU maintained its superiority with R$^2$ = 0.82, achieving statistically significant improvements over both baselines.
Notably, PI-CNN-BiGRU showed negligible improvement over PI-BiGRU-only in either metric, indicating that convolutional layers failed to capture meaningful inter-joint dependencies. 

These results validate that the MJCA module is crucial for modeling biomechanical coordination, with explicit multi-joint dependency modeling through attention providing substantial advantages over conventional approaches.

\begin{table}[t]
\centering
\caption{Loss ablation study with fixed backbone}
\label{tab:loss-ablation}
\footnotesize  
\setlength{\tabcolsep}{2.3pt}  
\begin{tabular}{ccc cc cc}
\toprule
\multicolumn{3}{c}{\textbf{Loss}} & \multicolumn{2}{c}{\textbf{Activation}} & \multicolumn{2}{c}{\textbf{Force}} \\
\cmidrule{1-3}\cmidrule{4-5}\cmidrule{6-7}
$\mathcal{L}_d$ & $\mathcal{L}_p$ & $\mathcal{L}_b$ & R$^2$  & NRMSE  & R$^2$  & NRMSE  \\
\midrule
\cmark & \cmark & \cmark & $\mathbf{0.84 \pm 0.15}$ & $\mathbf{0.08 \pm 0.04}$ & $\mathbf{0.82 \pm 0.18}$ & $\mathbf{0.09 \pm 0.04}$ \\
\cmark & \cmark & \xmark & $0.78 \pm 0.23$ & $0.10 \pm 0.04$ & $0.75 \pm 0.25$ & $0.12 \pm 0.04$ \\
\cmark & \xmark & \cmark & $0.65 \pm 0.33$ & $0.12 \pm 0.05$ & $0.63 \pm 0.16$ & $0.14 \pm 0.03$ \\
\xmark & \cmark & \cmark & $0.11 \pm 0.56$ & $0.22 \pm 0.06$ & $0.09 \pm 0.44$ & $0.22 \pm 0.05$ \\
\bottomrule
\end{tabular}
\end{table}

\section{Discussion}\label{discussions}

\subsection{Implications for Physics-Informed Learning}
Our PI-MJCA-BiGRU framework represents a middle path between traditional MSK modeling and purely data-driven methods. Multi-faceted experimental validation in Section. \ref{results} demonstrates that our physics-informed framework extracts representations highly similar to supervised methods while exhibiting superior generalization in cross-speed validation. This enhanced generalization capability indicates that physics-informed constraints help networks learn more fundamental biomechanical relationships rather than dataset-specific patterns.
Beyond accuracy metrics, physics-informed learning offers unique advantages for clinical translation. The label-free training enables deployment in settings where EMG collection is impractical—routine gait analysis, home rehabilitation monitoring, or resource-limited clinics. Models can be personalized using patient-specific skeletal geometry without sEMG calibration, adapting to pathological patterns or rehabilitation progress through physics constraints rather than labeled examples. The millisecond inference time, compared to hours for dynamic optimization, enables real-time integration with wearable sensors for continuous assessment.

\subsection{Limitations and Future Directions}
Despite the promising results, several limitations merit discussion. 
First, this study employed a simplified 2D planar model restricted to single-leg sagittal plane walking, whereas real-world locomotion involves 3D bilateral coordination, whole-body balance, and varied terrain adaptation. 
Extending to these scenarios introduces computational challenges through increased model complexity (multi-contact dynamics, terrain constraints) that may challenge the proposed physics-informed framework's stability. However, our success with multi-joint systems and biarticular muscles suggests feasibility for future whole-body implementations. 
Future work will focus on two main directions: (1) optimizing the network architecture to efficiently handle increased DOFs in 3D full lower-limb models, and (2) extending the embedded physics-informed constraints to incorporate complex locomotor dynamics beyond level walking, including stair climbing, slope walking, and other challenging scenarios.

A further limitation is the manual tuning of loss weights. The three loss terms currently require grid search to identify the balancing parameters $\omega_d$, $\omega_p$, and $\omega_b$, a process that is time-consuming and dataset-dependent. 
Future work will explore automated weighting strategies through uncertainty-based scaling, gradient balancing, or trainable parameters, reducing manual intervention while improving training stability for complex biomechanical models.
\section{Conclusion}\label{conclusion}
This paper presents a time-efficient, physics-informed deep learning framework for estimating muscle activations and forces directly from joint kinematics in multi-joint scenarios. The proposed MJCA-BiGRU architecture integrates joint cross-attention within a BiGRU to model inter-joint coordination in multi-joint systems. By embedding MSK modeling into the loss function, the framework ensures physiologically consistent predictions without large labeled datasets. Experiments on two datasets demonstrate accurate, time-efficient estimation of muscle states, highlighting its potential for clinical assessment, rehabilitation, and human–machine interaction.

\bibliographystyle{ieeetr}
\bibliography{reference}

@ARTICLE{TangTNSRE2021,
  author={Tang, Xiao and Zhang, Xu and Chen, Maoqi and Chen, Xiang and Chen, Xun},
  journal={IEEE Transactions on Neural Systems and Rehabilitation Engineering}, 
  title={Decoding Muscle Force From Motor Unit Firings Using Encoder-Decoder Networks}, 
  year={2021},
  volume={29},
  number={},
  pages={2484-2495},
  doi={10.1109/TNSRE.2021.3126752}}

@ARTICLE{HuTNSRE2022,
  author={Hu, Ruochen and Chen, Xiang and Zhang, Haotian and Zhang, Xu and Chen, Xun},
  journal={IEEE Transactions on Neural Systems and Rehabilitation Engineering}, 
  title={A Novel Myoelectric Control Scheme Supporting Synchronous Gesture Recognition and Muscle Force Estimation}, 
  year={2022},
  volume={30},
  number={},
  pages={1127-1137},
  doi={10.1109/TNSRE.2022.3166764}}

@ARTICLE{RobinTNSRE2023,
  author={Rubin, Noah and Hinson, Robert and Saul, Katherine and Hu, Xiaogang and Huang, He},
  journal={IEEE Transactions on Neural Systems and Rehabilitation Engineering}, 
  title={Ankle Torque Estimation With Motor Unit Discharges in Residual Muscles Following Lower-Limb Amputation}, 
  year={2023},
  volume={31},
  number={},
  pages={4821-4830},
  doi={10.1109/TNSRE.2023.3336543}}

@Article{Ao-JoNB-2024,
author={Ao, Di
and Fregly, Benjamin J.},
title={Comparison of synergy extrapolation and static optimization for estimating multiple unmeasured muscle activations during walking},
journal={Journal of NeuroEngineering and Rehabilitation},
year={2024},
month={Nov},
day={01},
volume={21},
number={1},
pages={194},

issn={1743-0003},
doi={10.1186/s12984-024-01490-y},
url={https://doi.org/10.1186/s12984-024-01490-y}
}

@article{zajac-CRiBE-1989,
  TITLE = {{Muscle and tendon: properties, models, scaling, and application to biomechanics and motor control}},
  AUTHOR = {Zajac, Felix E},
  URL = {https://hal.science/hal-04849267},
  JOURNAL = {{Critical Reviews in Biomedical Engineering}},
  PUBLISHER = {{Begell House}},
  VOLUME = {17},
  NUMBER = {4},
  PAGES = {359-411},
  YEAR = {1989},
  MONTH = Dec,
  HAL_ID = {hal-04849267},
  HAL_VERSION = {v1},
}

@ARTICLE{Wang-TNSRE-2025,
  author={Wang, Xiaohan and Li, Weidong and Song, Rong and Ao, Di and Hu, Huijing and Li, Le},
  journal={IEEE Transactions on Neural Systems and Rehabilitation Engineering}, 
  title={Corticomuscular Coupling Alterations During Elbow Isometric Contraction Correlated With Clinical Scores: An fNIRS-sEMG Study in Stroke Survivors}, 
  year={2025},
  volume={33},
  number={},
  pages={696-704},
  doi={10.1109/TNSRE.2025.3535928}}

@ARTICLE{cao-Tmech-2025,
  author={Cao, Yu and Ma, Shuhao and Zhang, Mengshi and Li, Zijian and Liu, Jindong and Huang, Jian and Zhang, Zhi-Qiang},
  journal={IEEE/ASME Transactions on Mechatronics}, 
  title={Musculoskeletal Model-Based Adaptive Variable Impedance Control With Flexible Prescribed Performance for Rehabilitation Robots}, 
  year={2025},
  volume={},
  number={},
  pages={1-9},
  doi={10.1109/TMECH.2025.3562670}}

@ARTICLE{Makaram-TIM-2021,
  author={Makaram, Navaneethakrishna and Karthick, P. A. and Swaminathan, Ramakrishnan},
  journal={IEEE Transactions on Instrumentation and Measurement}, 
  title={Analysis of Dynamics of EMG Signal Variations in Fatiguing Contractions of Muscles Using Transition Network Approach}, 
  year={2021},
  volume={70},
  number={},
  pages={1-8},
  doi={10.1109/TIM.2021.3063777}}

@article{Anderson-JoBE-2001,
    author = {Anderson, Frank C.  and Pandy, Marcus G. },
    title = {Dynamic Optimization of Human Walking },
    journal = {Journal of Biomechanical Engineering},
    volume = {123},
    number = {5},
    pages = {381-390},
    year = {2001},
    month = {05},
    issn = {0148-0731},
    doi = {10.1115/1.1392310},
    url = {https://doi.org/10.1115/1.1392310},
    eprint = {https://asmedigitalcollection.asme.org/biomechanical/article-pdf/123/5/381/5590682/381\_1.pdf},
}

@article{HAMNER-JoB-2010,
title = {Muscle contributions to propulsion and support during running},
journal = {Journal of Biomechanics},
volume = {43},
number = {14},
pages = {2709-2716},
year = {2010},
issn = {0021-9290},
doi = {https://doi.org/10.1016/j.jbiomech.2010.06.025},
url = {https://www.sciencedirect.com/science/article/pii/S0021929010003611},
author = {Samuel R. Hamner and Ajay Seth and Scott L. Delp}
}

@article{VALERO-JoB-2015,
title = {Exploring the high-dimensional structure of muscle redundancy via subject-specific and generic musculoskeletal models},
journal = {Journal of Biomechanics},
volume = {48},
number = {11},
pages = {2887-2896},
year = {2015},
note = {Symposia organized by the American Society of Biomechanics at the 7th World Congress of Biomechanics},
issn = {0021-9290},
doi = {https://doi.org/10.1016/j.jbiomech.2015.04.026},
url = {https://www.sciencedirect.com/science/article/pii/S0021929015002468},
author = {F.J. Valero-Cuevas and B.A. Cohn and H.F. Yngvason and E.L. Lawrence}
}

@article{Sobinov-PLOs-2020,
    doi = {10.1371/journal.pcbi.1008350},
    author = {Sobinov, Anton AND Boots, Matthew T. AND Gritsenko, Valeriya AND Fisher, Lee E. AND Gaunt, Robert A. AND Yakovenko, Sergiy},
    journal = {PLOS Computational Biology},
    publisher = {Public Library of Science},
    title = {Approximating complex musculoskeletal biomechanics using multidimensional autogenerating polynomials},
    year = {2020},
    month = {12},
    volume = {16},
    url = {https://doi.org/10.1371/journal.pcbi.1008350},
    pages = {1-26},
    number = {12},}

@Article{Schmidt-BEO-2023,
author={Schmidt, Marie D.
and Glasmachers, Tobias
and Iossifidis, Ioannis},
title={The concepts of muscle activity generation driven by upper limb kinematics},
journal={BioMedical Engineering OnLine},
year={2023},
month={Jun},
day={24},
volume={22},
number={1},
pages={63},
issn={1475-925X},
doi={10.1186/s12938-023-01116-9},
url={https://doi.org/10.1186/s12938-023-01116-9}}

@ARTICLE{Nasr-FiCN-2021,
AUTHOR={Nasr, Ali  and Inkol, Keaton A.  and Bell, Sydney  and McPhee, John },       
TITLE={InverseMuscleNET: Alternative Machine Learning Solution to Static Optimization and Inverse Muscle Modeling},     
JOURNAL={Frontiers in Computational Neuroscience},     
VOLUME={Volume 15 - 2021},
YEAR={2021},
URL={https://www.frontiersin.org/journals/computational-neuroscience/articles/10.3389/fncom.2021.759489},
DOI={10.3389/fncom.2021.759489},
ISSN={1662-5188},}

@Article{khant-MDPIS-2023,
AUTHOR = {Khant, Min and Gouwanda, Darwin and Gopalai, Alpha A. and Lim, King Hann and Foong, Chee Choong},
TITLE = {Estimation of Lower Extremity Muscle Activity in Gait Using the Wearable Inertial Measurement Units and Neural Network},
JOURNAL = {Sensors},
VOLUME = {23},
YEAR = {2023},
NUMBER = {1},
ARTICLE-NUMBER = {556},
URL = {https://www.mdpi.com/1424-8220/23/1/556},
PubMedID = {36617154},
ISSN = {1424-8220},
DOI = {10.3390/s23010556}
}

@ARTICLE{Cohn-FiRS-2023,
AUTHOR={Cohn, Brian A.  and Valero-Cuevas, Francisco J. },     
TITLE={Muscle redundancy is greatly reduced by the spatiotemporal nature of neuromuscular control},     
JOURNAL={Frontiers in Rehabilitation Sciences},      
VOLUME={Volume 4 - 2023},
YEAR={2023},
URL={https://www.frontiersin.org/journals/rehabilitation-sciences/articles/10.3389/fresc.2023.1248269},
DOI={10.3389/fresc.2023.1248269},
ISSN={2673-6861},}

@article{TRINLER-JoB-2019,
title = {Muscle force estimation in clinical gait analysis using AnyBody and OpenSim},
journal = {Journal of Biomechanics},
volume = {86},
pages = {55-63},
year = {2019},
issn = {0021-9290},
doi = {https://doi.org/10.1016/j.jbiomech.2019.01.045},
url = {https://www.sciencedirect.com/science/article/pii/S002192901930082X},
author = {Ursula Trinler and Hermann Schwameder and Richard Baker and Nathalie Alexander},}

@article{Kannape-JoN-2013,
author = {Kannape, O. A. and Blanke, O.},
title = {Self in motion: sensorimotor and cognitive mechanisms in gait agency},
journal = {Journal of Neurophysiology},
volume = {110},
number = {8},
pages = {1837-1847},
year = {2013},
doi = {10.1152/jn.01042.2012},
note ={PMID: 23825398},
URL = {https://doi.org/10.1152/jn.01042.2012},
eprint = {https://doi.org/10.1152/jn.01042.2012},}

@Article{Rane-AoBE-2019,
author={Rane, Lance
and Ding, Ziyun
and McGregor, Alison H.
and Bull, Anthony M. J.},
title={Deep Learning for Musculoskeletal Force Prediction},
journal={Annals of Biomedical Engineering},
year={2019},
month={Mar},
day={01},
volume={47},
number={3},
pages={778-789},
issn={1573-9686},
doi={10.1007/s10439-018-02190-0},
url={https://doi.org/10.1007/s10439-018-02190-0},}

@Article{Horst-SD-2021,
author={Horst, Fabian
and Slijepcevic, Djordje
and Simak, Marvin
and Sch{\"o}llhorn, Wolfgang I.},
title={Gutenberg Gait Database, a ground reaction force database of level overground walking in healthy individuals},
journal={Scientific Data},
year={2021},
month={Sep},
day={02},
volume={8},
number={1},
pages={232},
issn={2052-4463},
doi={10.1038/s41597-021-01014-6},
url={https://doi.org/10.1038/s41597-021-01014-6},}

@Article{Karniadakis-NRP-2021,
author={Karniadakis, George Em
and Kevrekidis, Ioannis G.
and Lu, Lu
and Perdikaris, Paris
and Wang, Sifan
and Yang, Liu},
title={Physics-informed machine learning},
journal={Nature Reviews Physics},
year={2021},
month={Jun},
day={01},
volume={3},
number={6},
pages={422-440},
issn={2522-5820},
doi={10.1038/s42254-021-00314-5},
url={https://doi.org/10.1038/s42254-021-00314-5}
}

@ARTICLE{Falisse-TBME-2025,
  author={Falisse, Antoine and Uhlrich, Scott D. and Chaudhari, Akshay S. and Hicks, Jennifer L. and Delp, Scott L.},
  journal={IEEE Transactions on Biomedical Engineering}, 
  title={Marker Data Enhancement for Markerless Motion Capture}, 
  year={2025},
  volume={72},
  number={6},
  pages={2013-2022},
  doi={10.1109/TBME.2025.3530848}}

@ARTICLE{Ma-TNSRE-2025,
  author={Ma, Shuhao and Cao, Yu and Robertson, Ian D. and Shi, Chaoyang and Liu, Jindong and Zhang, Zhi-Qiang},
  journal={IEEE Transactions on Neural Systems and Rehabilitation Engineering}, 
  title={Knowledge-Based Deep Learning for Time-Efficient Inverse Dynamics}, 
  year={2025},
  volume={33},
  number={},
  pages={522-531},
  doi={10.1109/TNSRE.2025.3530992}}

@ARTICLE{Zhang-TNSRE-2023,
  author={Zhang, Jie and Zhao, Yihui and Shone, Fergus and Li, Zhenhong and Frangi, Alejandro F. and Xie, Sheng Quan and Zhang, Zhi-Qiang},
  journal={IEEE Transactions on Neural Systems and Rehabilitation Engineering}, 
  title={Physics-Informed Deep Learning for Musculoskeletal Modeling: Predicting Muscle Forces and Joint Kinematics From Surface EMG}, 
  year={2023},
  volume={31},
  number={},
  pages={484-493},
  doi={10.1109/TNSRE.2022.3226860}}

@ARTICLE{Zhang-JBHI-2022,
  author={Zhang, Longbin and Zhu, Xueyu and Gutierrez-Farewik, Elena M. and Wang, Ruoli},
  journal={IEEE Journal of Biomedical and Health Informatics}, 
  title={Ankle Joint Torque Prediction Using an NMS Solver Informed-ANN Model and Transfer Learning}, 
  year={2022},
  volume={26},
  number={12},
  pages={5895-5906},
  doi={10.1109/JBHI.2022.3207313}}

@ARTICLE{Han-TIM-2025,
  author={Han, Lijun and Cheng, Long and Zou, Yongxiang and Li, Yanan},
  journal={IEEE Transactions on Instrumentation and Measurement}, 
  title={Physics-Informed Deep Transfer Learning for sEMG-Based Multiple Joint Angle and Torque Estimation}, 
  year={2025},
  volume={74},
  number={},
  pages={1-13},
  doi={10.1109/TIM.2025.3572159}}

@article{Seth-PLOSCB-2018,
    doi = {10.1371/journal.pcbi.1006223},
    author = {Seth, Ajay AND Hicks, Jennifer L. AND Uchida, Thomas K. AND Habib, Ayman AND Dembia, Christopher L. AND Dunne, James J. AND Ong, Carmichael F. AND DeMers, Matthew S. AND Rajagopal, Apoorva AND Millard, Matthew AND Hamner, Samuel R. AND Arnold, Edith M. AND Yong, Jennifer R. AND Lakshmikanth, Shrinidhi K. AND Sherman, Michael A. AND Ku, Joy P. AND Delp, Scott L.},
    journal = {PLOS Computational Biology},
    publisher = {Public Library of Science},
    title = {{OpenSim}: Simulating musculoskeletal dynamics and neuromuscular control to study human and animal movement},
    year = {2018},
    volume = {14},
    url = {https://doi.org/10.1371/journal.pcbi.1006223},
    pages = {1-20}}

@article{CAMARGO2021110320,
title = {A comprehensive, open-source dataset of lower limb biomechanics in multiple conditions of stairs, ramps, and level-ground ambulation and transitions},
journal = {Journal of Biomechanics},
volume = {119},
pages = {110320},
year = {2021},
issn = {0021-9290},
doi = {https://doi.org/10.1016/j.jbiomech.2021.110320},
url = {https://www.sciencedirect.com/science/article/pii/S0021929021001007},
author = {Jonathan Camargo and Aditya Ramanathan and Will Flanagan and Aaron Young}
}

@ARTICLE{wei-jbhi-2025,
  author={Wei, Zijun and Zhang, Zhiqiang and Xie, Sheng Quan},
  journal={IEEE Journal of Biomedical and Health Informatics}, 
  title={A Transformer Framework Informed by Muscle Anatomy and Sequence-to-Sequence Translation for Continuous Joint Kinematics Prediction Using sEMG}, 
  year={2025},
  volume={},
  number={},
  pages={1-13},
  doi={10.1109/JBHI.2025.3589889}}

@ARTICLE{lin-TNSRE-2023,
  author={Lin, Chuang and Chen, Xingjian and Guo, Weiyu and Jiang, Ning and Farina, Dario and Su, Jingyong},
  journal={IEEE Transactions on Neural Systems and Rehabilitation Engineering}, 
  title={A BERT Based Method for Continuous Estimation of Cross-Subject Hand Kinematics From Surface Electromyographic Signals}, 
  year={2023},
  volume={31},
  number={},
  pages={87-96},
  doi={10.1109/TNSRE.2022.3216528}}

@ARTICLE{Schonhaut-TBME-2025,
  author={Schonhaut, Ethan B. and Scherpereel, Keaton L. and Young, Aaron J.},
  journal={IEEE Transactions on Biomedical Engineering}, 
  title={Is EMG Information Necessary for Deep Learning Estimation of Joint and Muscle Level States?}, 
  year={2025},
  volume={},
  number={},
  pages={1-12},
  doi={10.1109/TBME.2025.3577084}}

@ARTICLE{Dubois-JoNB-2023,
  title    = "A guide to inter-joint coordination characterization for discrete
              movements: a comparative study",
  author   = "Dubois, Oc{\'e}ane and Roby-Brami, Agn{\`e}s and Parry, Ross and
              Khoramshahi, Mahdi and Jarrass{\'e}, Nathana{\"e}l",
  journal  = "Journal of NeuroEngineering and Rehabilitation",
  volume   =  20,
  number   =  1,
  pages    = "132",
  month    =  sep,
  year     =  2023
}

@ARTICLE{Zhang-TASE-2025,
  author={Zhang, Bi and Jiang, Weizhong and Tan, Xiaowei and Su, Juhua and Zhao, Yiwen and Zhao, Xingang},
  journal={IEEE Transactions on Automation Science and Engineering}, 
  title={Modular Soft Exoskeleton Design and Control for Assisting Movements in Multiple Lower Limb Joint Configurations}, 
  year={2025},
  volume={22},
  number={},
  pages={20799-20813},
  doi={10.1109/TASE.2025.3607351}}

@ARTICLE{Liang-TNSRE-2023,
  author={Liang, Tie and Miao, Huacong and Wang, Hongrui and Liu, Xiaoguang and Liu, Xiuling},
  journal={IEEE Transactions on Neural Systems and Rehabilitation Engineering}, 
  title={Surface Electromyography-Based Analysis of the Lower Limb Muscle Network and Muscle Synergies at Various Gait Speeds}, 
  year={2023},
  volume={31},
  number={},
  pages={1230-1237},
  doi={10.1109/TNSRE.2023.3242911}}

@ARTICLE{Collings-MSSE-2023,
  title    = "Gluteal Muscle Forces during {Hip-Focused} Injury Prevention and
              Rehabilitation Exercises",
  author   = "Collings, Tyler J and Bourne, Matthew N and Barrett, Rod S and
              Meinders, Evy and GON{\c c}ALVES, Bas{\'\i}lio A M and Shield,
              Anthony J and Diamond, Laura E",
  journal  = "Medicine \& Science in Sports \& Exercise",
  volume   =  55,
  number   =  4,
  year     =  2023
}

@ARTICLE{Li-TNSRE-2022,
  author={Li, Quanlin and Xia, Yang and Wang, Xianzong and Xin, Peiyang and Chen, Wenbin and Xiong, Caihua},
  journal={IEEE Transactions on Neural Systems and Rehabilitation Engineering}, 
  title={Muscle-Effort-Minimization-Inspired Kinematic Redundancy Resolution for Replicating Natural Posture of Human Arm}, 
  year={2022},
  volume={30},
  number={},
  pages={2341-2351},
  doi={10.1109/TNSRE.2022.3198400}}

@ARTICLE{Cao-TRO-2025,
  author={Cao, Yu and Zhang, Mengshi and Huang, Jian and Mohammed, Samer},
  journal={IEEE Transactions on Robotics}, 
  title={Load-Transfer Suspended Backpack With Bioinspired Vibration Isolation for Shoulder Pressure Reduction Across Diverse Terrains}, 
  year={2025},
  volume={41},
  number={},
  pages={3059-3077},
  doi={10.1109/TRO.2025.3562488}}

\vspace{12pt}
\color{red}

\end{document}